\pdfoutput=1

\documentclass[11pt]{article}
\usepackage{authblk}
\usepackage{ACL2023}

\usepackage{times}
\usepackage{latexsym}

\usepackage[T1]{fontenc}

\usepackage[utf8]{inputenc}

\usepackage{microtype}

\usepackage{inconsolata}
\usepackage{booktabs}
\usepackage{multirow}
\usepackage{amsmath, amssymb, amsthm, amscd, amsfonts}
\usepackage{IEEEtrantools}
\usepackage{times}
\usepackage{soul}
\usepackage{url}
\usepackage[utf8]{inputenc}
\usepackage{graphicx}
\usepackage{amsmath}
\usepackage{amsthm}
\usepackage{booktabs}
\usepackage{algorithm}
\usepackage{algorithmic}
\usepackage{natbib} 
\usepackage{amssymb}
\usepackage{amsfonts} 
\usepackage{multirow}
\usepackage{boldline}
\usepackage{hhline}
\usepackage{amsmath}
\usepackage{xcolor}
\usepackage{pifont}
\usepackage{subfigure}
\usepackage{makecell}
\usepackage{utfsym}
\usepackage{bbding}
\usepackage[normalem]{ulem}
\useunder{\uline}{\ul}{}
\usepackage[inline]{enumitem}
\usepackage{enumitem}
\usepackage{IEEEtrantools}
\usepackage{stmaryrd}

\theoremstyle{definition}

\theoremstyle{remark}

\title{\texttt{RethinkingTMSC}: An Empirical Study for Target-Oriented Multimodal Sentiment Classification}

\author{
    \bf{\normalsize
    Junjie Ye$^{1}$, \ \ Jie Zhou$^{2}$\thanks{$^*$  Corresponding authors, jzhou@cs.ecnu.edu.cn, qz@fudan.edu.cn.} , \ \ Junfeng Tian$^{3}$, \ \ Rui Wang$^{3}$,} \\
    \bf{\normalsize Qi Zhang$^{1*}$, \ \ Tao Gui$^{4}$, \ \ Xuanjing Huang$^1$} \\ 
  {$^1$ \normalsize School of Computer Science, Fudan University} \\
  {$^2$ \normalsize School of Computer Science and Technology, East China Normal University} \\
  {$^3$ \normalsize nyonic, Shanghai, China} \\
  {$^4$ \normalsize Institute of Modern Languages and Linguistics, Fudan University} \\
  }

\begin{document}
\maketitle
\begin{abstract}

Recently, Target-oriented Multimodal Sentiment Classification (TMSC) has gained significant attention among scholars. However, current multimodal models have reached a performance bottleneck.
To investigate the causes of this problem, we perform extensive empirical evaluation and in-depth analysis of the datasets to answer the following questions:
\textbf{Q1}: Are the modalities equally important for TMSC? 
\textbf{Q2}: Which multimodal fusion modules are more effective? 
\textbf{Q3}: Do existing datasets adequately support the research?
Our experiments and analyses reveal that the current TMSC systems primarily rely on the textual modality, as most of targets' sentiments can be determined \emph{solely} by text.
{Consequently, we point out several directions to work on for the TMSC task in terms of model design and dataset construction.}
The code and data can be found in \url{https://github.com/Junjie-Ye/RethinkingTMSC}.

\end{abstract}

\section{Introduction}
\label{sect:intro}
\begin{table*}[t!]
    \centering
    \small
    \resizebox{\textwidth}{!}{
    \begin{tabular}{l|ccc|ccccc}
    \toprule
        \multirow{2}*{Model}    & \multicolumn{3}{c|}{Image Encoder}    & \multicolumn{5}{c}{Fusion Module}  \\
        ~ & ResNet & ViT & Faster R-CNN & Concat & Tensor Fusion & Self Attention & Image2Text  & Text2Image \\\midrule 
        Res-BERT+BL  & \Checkmark & \XSolidBrush & \XSolidBrush & \Checkmark & \XSolidBrush & \Checkmark &  \XSolidBrush  &  \XSolidBrush \\
        Res-BERT+BL-TFN  & \Checkmark & \XSolidBrush & \XSolidBrush & \XSolidBrush & \Checkmark & \Checkmark &  \XSolidBrush  &  \XSolidBrush  \\
        mBERT   & \Checkmark & \XSolidBrush & \XSolidBrush & \Checkmark & \XSolidBrush &   \Checkmark & \XSolidBrush & \XSolidBrush \\
        TomBERT  & \Checkmark & \XSolidBrush & \XSolidBrush & \Checkmark & \XSolidBrush & \Checkmark & \Checkmark  & \XSolidBrush \\
        EF-CapTrBERT   & \Checkmark & \XSolidBrush & \XSolidBrush & \XSolidBrush & \XSolidBrush &  \Checkmark & \XSolidBrush  & \XSolidBrush \\
        SMP  & \XSolidBrush & \Checkmark & \XSolidBrush & \XSolidBrush & \XSolidBrush & \XSolidBrush & \Checkmark & \Checkmark \\
        VLP   & \XSolidBrush & \XSolidBrush & \Checkmark & \XSolidBrush & \XSolidBrush &  \Checkmark & \XSolidBrush & \XSolidBrush\\
        \bottomrule
    \end{tabular}
    }
    \caption{The model structures of various baselines. All text encoders in the above models except for VLP are initialized with BERT.}
    \label{baseline}
    \vspace{-4mm}
\end{table*}

Target-oriented sentiment classification, also known as aspect-based sentiment classification, is a fundamental task of sentiment analysis \cite{manandhar2014semeval,pontiki2015semeval,pontiki2016semeval}. It aims to judge the sentimental polarity (positive, negative, or neutral) of a specific target within text.
To improve the performance by considering multimodal information, Target-oriented Multimodal Sentiment Classification (TMSC) is proposed to integrate both visual and textual information \cite{yu2019adapting}. 

\begin{figure}[t!]
\begin{center}
\includegraphics[width=0.48\textwidth]{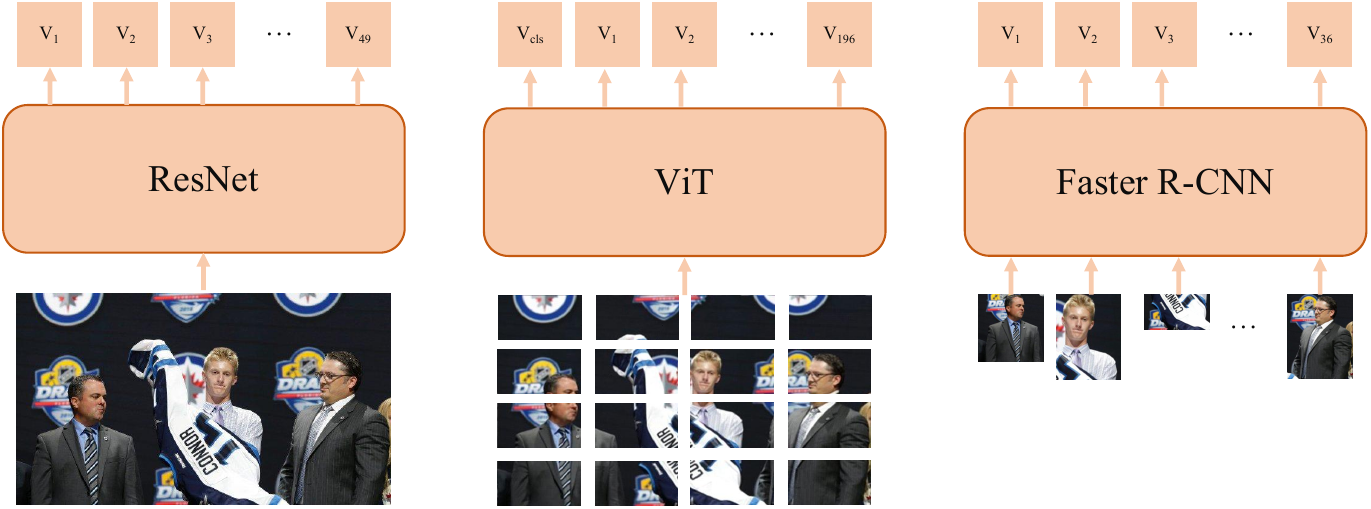}
\end{center}
\vspace{-2mm}
\caption{Different image encoders. } 
\label{Encoders}
\vspace{-4mm}
\end{figure}

Recently, the performance of the TMSC systems gradually reaches a plateau and the progress in tackling this task has slowed down.
Using the F1-score metric on the popular datasets, Twitter15 and Twitter17 \cite{yu2019adapting}, we observe that state-of-the-art baselines only achieve an F1-score of around 70.
Therefore, in this paper, we aim to analyze the causes behind it at both model level and modality level.
Roughly speaking, the modules in the model structures can be categorized into two types: 1) encoders to model the representations of different modalities; and 2) multimodal fusion modules to model the interactions between modalities. 
Moreover, we give a deep analysis of the characteristics of two widely-used datasets, aiming to answer the following three questions:

\textbf{Q1}: Are the modalities equally important for TMSC?
To explore this issue, we compare and analyze the performance of unimodal models on this task.
For the textual modality, we use BERT \cite{DBLP:conf/naacl/DevlinCLT19} as the backbone, as it is a widely-used pre-trained language model outperforming earlier models like LSTM \cite{hochreiter1997long}, memory network \cite{weston2014memory}, etc. 
For the visual modality, ResNet \cite{2016Deep}, ViT \cite{DBLP:conf/iclr/DosovitskiyB0WZ21}, and Faster R-CNN \cite{ren2015faster} are adopted (see Figure~\ref{Encoders}).

\textbf{Q2}: Which multimodal fusion modules are more effective?
The current models use various fusion strategies to model the interactions between modalities, while obtaining little improvement. 
To explore the effectiveness of different fusion approaches,
we summarize the fusion strategies into six categories: Concatenation, Tensor Fusion \cite{2017Tensor}, Self Attention, 
Image2Text, Text2Image and Bi-direction.
Then we perform a comparative study of 
them using a unified setup to eliminate potential bias from model size and structure (see Figure~\ref{Fusion Modules}).

\textbf{Q3}: Do existing datasets adequately support the research?
We analyze the existing datasets (i.e., Twitter15 and Twitter17) in depth 
and obtain the following findings:
1) The size of existing datasets is limited; 
2) The multimodal sentiment is much more consistent with the textual sentiment than the visual sentiment;
3) A large number of targets do not exist in images;
4) There are only a small number of samples where the sentiment is decided by both text and image.

The main contributions of this work are as follows: 1) We investigate the effectiveness of different model structures for TMSC, including various unimodal encoders and multimodal fusion modules; 2) We give an in-depth analysis of limitations of existing widely-used datasets; 
3) We derive several valuable observations and point out promising directions for the future research of TMSC model design and dataset creation.

\section{Empirical Study}
\label{sect:Empirical Study}

We summarize the model structures and performance of the baselines for the TMSC task in Table~\ref{baseline}. Their structural differences are mainly reflected in the different unimodal encoders and multimodal fusion modules used. Therefore, we carry out several experiments to analyze the impact of these two aspects on performance.

\subsection{Unimodal Encoder}
As previously mentioned in Section~\ref{sect:intro}, we primarily focus on exploring the different image encoders, ResNet, ViT, and Faster R-CNN (see Figure~\ref{Encoders}),
while using BERT as the text encoder.

\begin{figure*}[t!]
\begin{center}
\includegraphics[width=0.9\textwidth]{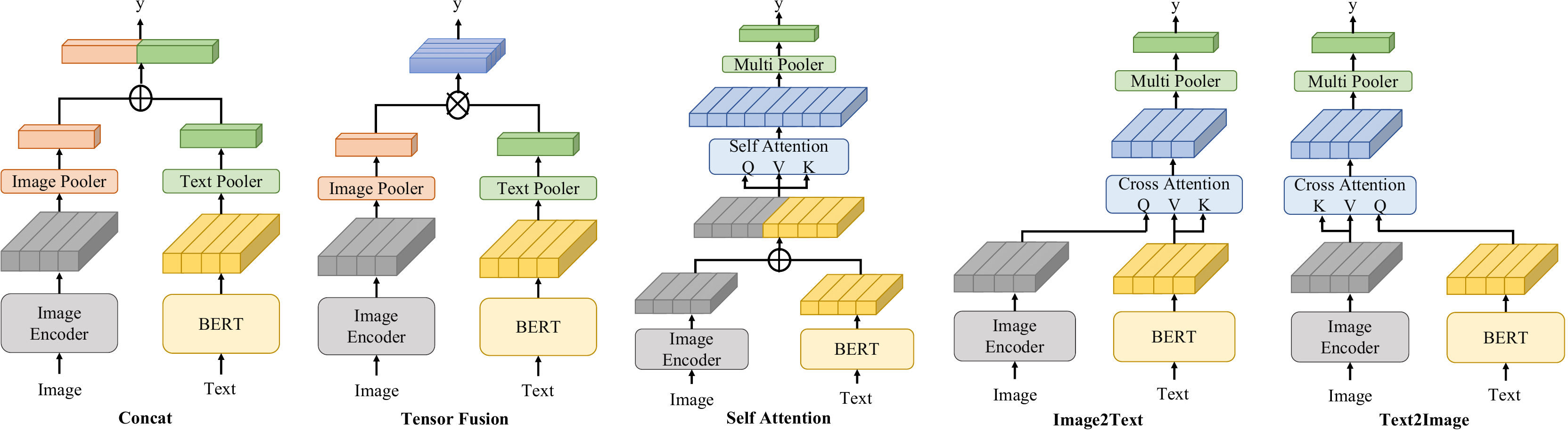}
\end{center}
\vspace{-3mm}
\caption{Various multimodal fusion modules. } 
\label{Fusion Modules}
\vspace{-3mm}
\end{figure*}

\textbf{ResNet.} Following most of the baselines (e.g., mBERT~\cite{yu2019adapting}, TomBERT~\cite{yu2019adapting} and EF-CapTrBERT~\cite{2021Exploiting}), we adopt ResNet-152 as one of the image encoders.
Each image is resized into 224 by 224, and then passed through the model to obtain 49 regions, which are used as the image representation $I=[v_1, v_2, ..., v_{49}]$, where $v_i \in \mathbb{R}^{2048}$.

\textbf{ViT.} Following SMP~\cite{ye2022sentiment}, we adopt ViT to model the image by dividing it into 16 by 16 patches. 
A CLS token is added at the beginning and fed into the Transformer~\cite{DBLP:conf/nips/VaswaniSPUJGKP17} encoder to obtain the image representation $I=[v_{cls}, v_1, v_2, ..., v_{196}]$, where $v_i \in \mathbb{R}^{768}$.

\textbf{Faster R-CNN.} 
Similar to VLP~\cite{DBLP:conf/acl/LingYX22}, we adopt Faster R-CNN that is retrained on the Visual Genome dataset \citep{krishna2017visual}. 
We select the top $36$ object proposals as the image representation $I=[v_1, v_2, ..., v_{36}]$,
where $v_i \in \mathbb{R}^{2048}$ is obtained from the ROI pooling layer of the Region Proposal Network~\cite{ren2015faster}.

\subsection{Multimodal Fusion}
We categorize the current multimodal fusion modules into six groups as follows (see Figure~\ref{Fusion Modules}).

\textbf{Concatenate} is the simplest form of fusion, where the pooled text representation $H^T_p \in \mathbb{R}^{768}$ is directly combined with the pooled image representation $H^I_p \in \mathbb{R}^{768}$~\footnote{A linear mapping layer is added after the image encoder to map the image representation to 768 dimensions to ensure uniformity when using different image encoders.} to obtain the multimodal representation $H=H^I_p \bigoplus H^T_p$, where $\bigoplus$ is a concatenation operation and $H \in \mathbb{R}^{768+768}$. 

\textbf{Tensor Fusion} is proposed for modeling interactions between modalities while preserving the characteristics of individual modalities.
We obtain 
$H=H^I_p \bigotimes H^T_p$,
where $\bigotimes$ is an outer product operation and $H \in \mathbb{R}^{768 \times 768}$.

\textbf{Self Attention} concatenates the image representation $H^I \in \mathbb{R}^{l_I \times 768}$ and the text representation $H^T \in \mathbb{R}^{l_T \times 768}$, 
where $l_I$ and $l_T$ are the lengths of image and text, respectively. Then it is passed through three self-attention layers and a pooling layer to obtain 
$H \in \mathbb{R}^{768}$.

\textbf{Image2Text} is one type of cross-attention mechanism, using 
$H^I$ as the query and 
$H^T$ as the key and value, through three attention layers to get 
$H \in \mathbb{R}^{768}$. \textbf{Text2Image} uses $H^T$ as the query and 
$H^I$ as the key and value instead. Furthermore, we concatenate these two as \textbf{Bi-direction} representation $H \in \mathbb{R}^{768+768}$.


\begin{table}[t!]
    \centering
    \scriptsize
    \setlength{\tabcolsep}{0.6mm}{\begin{tabular}{llcccc}
    \toprule
        \multirow{2}*{Modality} & \multirow{2}*{Model} & \multicolumn{2}{c}{Twitter15} & \multicolumn{2}{c}{Twitter17}   \\ 
        ~ & ~ & ACC & F1 & ACC & F1  \\\midrule 
        Text & BERT & 76.72±1.16 & 71.19±2.19 & 68.04±0.40 & 65.66±0.35  \\ \midrule 
        \multirow{3}*{Image} & ResNet & 57.65±1.00 & 32.52±2.66 & 57.79±0.99 & 51.98±1.23  \\         
        ~ & ViT & \underline{59.65±1.13} & 31.25±2.71 & \underline{59.53±0.95} & \underline{54.08±0.78}  \\
        ~ & Faster R-CNN & 55.97±1.10 & \underline{35.72±5.43} & 56.18±0.85 & 49.88±1.70  \\  \midrule 
        \multirow{21}*{Multimodal} 
        ~ & \multicolumn{5}{c}{ResNet} \\ \cmidrule{2-6}
        ~ & Concatenate & 75.29±0.45 & 68.71±1.34 & 67.92±0.56 & 65.32±0.53  \\ 
        ~ & Tensor Fusion & 74.19±0.94 & 68.93±0.57 & 66.66±1.21 & 63.99±1.61  \\ 
        ~ & Self Attention & 76.03±0.96 & 70.57±2.39 & 68.01±0.96 & 65.41±1.60  \\ 
        ~ & Image2Text & 77.13±1.33 & 71.48±1.90 & \underline{69.37±0.36} & \underline{66.85±0.79}  \\ 
        ~ & Text2Image & 75.18±1.66 & 67.77±4.81 & 68.07±0.58 & 65.18±1.48  \\ 
        ~ & Bi-direction & \underline{77.32±0.63} & \underline{\textbf{72.06±0.81}} & 68.41±1.01 & 66.39±1.39  \\ 
        \cmidrule{2-6}
        & \multicolumn{5}{c}{ViT} \\ \cmidrule{2-6} 
        ~ & Concatenate & 76.22±0.90 & 70.37±1.45 & 67.94±0.70 & 66.17±0.78  \\ 
        ~ & Tensor Fusion & 73.44±0.78 & 67.46±1.45 & 65.46±1.67 & 62.02±1.40  \\ 
        ~ & Self Attention & 75.08±0.41 & 68.94±0.83 & 67.52±0.58 & 65.56±0.35  \\
        ~ & Image2Text & \underline{77.11±0.44} & \underline{71.91±0.42} & 69.14±0.52 & 66.96±0.68  \\ 
        ~ & Text2Image & 75.12±1.01 & 69.40±1.38 & 67.52±1.06 & 64.49±1.46  \\ 
        ~ & Bi-direction & 76.70±0.75 & 71.67±1.45 & \underline{69.16±0.17} & \underline{67.25±0.56}  \\  
        \cmidrule{2-6}
        & \multicolumn{5}{c}{Faster R-CNN} \\ \cmidrule{2-6}
        ~ & Concatenate & 75.45±0.73 & 69.77±1.23 & 67.60±1.15 & 64.74±1.69  \\ 
        ~ & Tensor Fusion & 72.09±0.66 & 66.77±1.04 & 66.34±1.45 & 62.96±2.09   \\ 
        ~ & Self Attention & 76.09±0.89 & 70.08±1.37 & 68.09±1.10 & 66.12±1.23   \\ 
        ~ & Image2Text & \underline{\textbf{77.36±0.37}} & \underline{71.69±0.37} & 68.43±0.65 & 66.44±1.10  \\ 
        ~ & Text2Image & 70.82±2.99 & 57.94±5.81 & 60.31±6.43 & 54.50±7.06   \\ 
        ~ & Bi-direction & 76.57±0.46 & 70.88±0.89 & \underline{\textbf{69.51±0.62}} & \underline{\textbf{67.50±0.37}}  \\  \bottomrule  
    \end{tabular}}
    \caption{Results on Twitter15 and Twitter17. The overall best results and those within each corresponding block are marked with \textbf{bold} and \underline{underline}, respectively.}
    \label{experiment}
    \vspace{-4mm}
\end{table}

\begin{table*}[t!]
\centering
\small
\resizebox{0.90\textwidth}{!}{\begin{tabular}{l|ccccc|ccccc}
\toprule
 \multirow{2}*{Dataset}       & \multicolumn{5}{c|}{Twitter15}     &        \multicolumn{5}{c}{Twitter17}           \\
   & \#Negative & \#Neutral & \#Positive & \#Total & \#Avg Targets & \#Negative & \#Neutral & \#Positive & \#Total & \#Avg Targets \\ \midrule
Train                    & 368      & 1883    & 928      & 3179  & 1.348   & 416 & 1638 & 1508 & 3562  & 1.410  \\
Dev                      & 149      & 670     & 303      & 1122  & 1.336    & 144 & 517 & 515 & 1176  & 1.439  \\
Test                     & 113      & 607     & 317      & 1037  & 1.354   & 168 & 573 & 493 & 1234 & 1.450   \\
\bottomrule 
\end{tabular}}
\caption{Statistics of the datasets. \#Avg Targets means the average number of targets for each sample.}
\label{dataset}
\vspace{-3mm}
\end{table*}

\subsection{Results Analysis}
We perform experiments of different unimodal encoders and fusion modules over Twitter15 and Twitter17. 
In Table~\ref{experiment}, we show the results and we have the following observations{~\footnote{The experimental setup is illustrated in Appendix \ref{setup}.}}:

\textbf{First}, the text-only model (i.e., BERT) performs well, while the visual-only models (i.e., ResNet, ViT, and Faster R-CNN) perform relatively poorly, revealing that the reliance on text is much greater than that on images for the TMSC task on these two datasets. In comparison, this phenomenon is more pronounced in Twitter15.

\textbf{Second}, the performance of the model is affected by the use of different fusion methods. Specifically, fusion modules that primarily focus on acquiring the textual information (e.g., Image2Text) perform better than those focused on acquiring the visual information (e.g., Text2Image). This again reveals the inconsistent importance of text and images.

\textbf{Third}, compared with the text-only model, the various fusion modules do not have significant gains in performance and some are even worse. This is due to the fact that some images do not provide related information, but rather distracting information instead~\footnote{We give a detailed analysis of the performance comparison for the multimodal model versus the text-only model in Appendix~\ref{visualization}.}.


\textbf{Fourth}, the impact of various image encoders is not clear, as evidenced by low performance and high standard deviation on the two datasets (see the ``Image'' part of Table~\ref{experiment}). Moreover, differences in performance among various image encoders are small in the multimodal fusion settings (see the ``Multimodal'' part of Table~\ref{experiment}). 
This is due to the characteristics of visual data in existing datasets, 
which is analyzed in depth in the following section.

{Based on the comprehensive experimental analyses conducted above, we identify several key points to be considered when designing models for the TMSC task in the future: 1) leveraging text information to exploit the advantages of textual data fully; 2) devising more effective image encoding methods to extract semantic information from images better; and 3) enhancing the noise immunity of the fusion module to enable more flexible selection and utilization of informative features from both textual and visual modalities.}



\section{Data Analysis}
\label{sec:data}

\begin{figure}[t!]
\begin{center}
\includegraphics[width=0.45\textwidth]{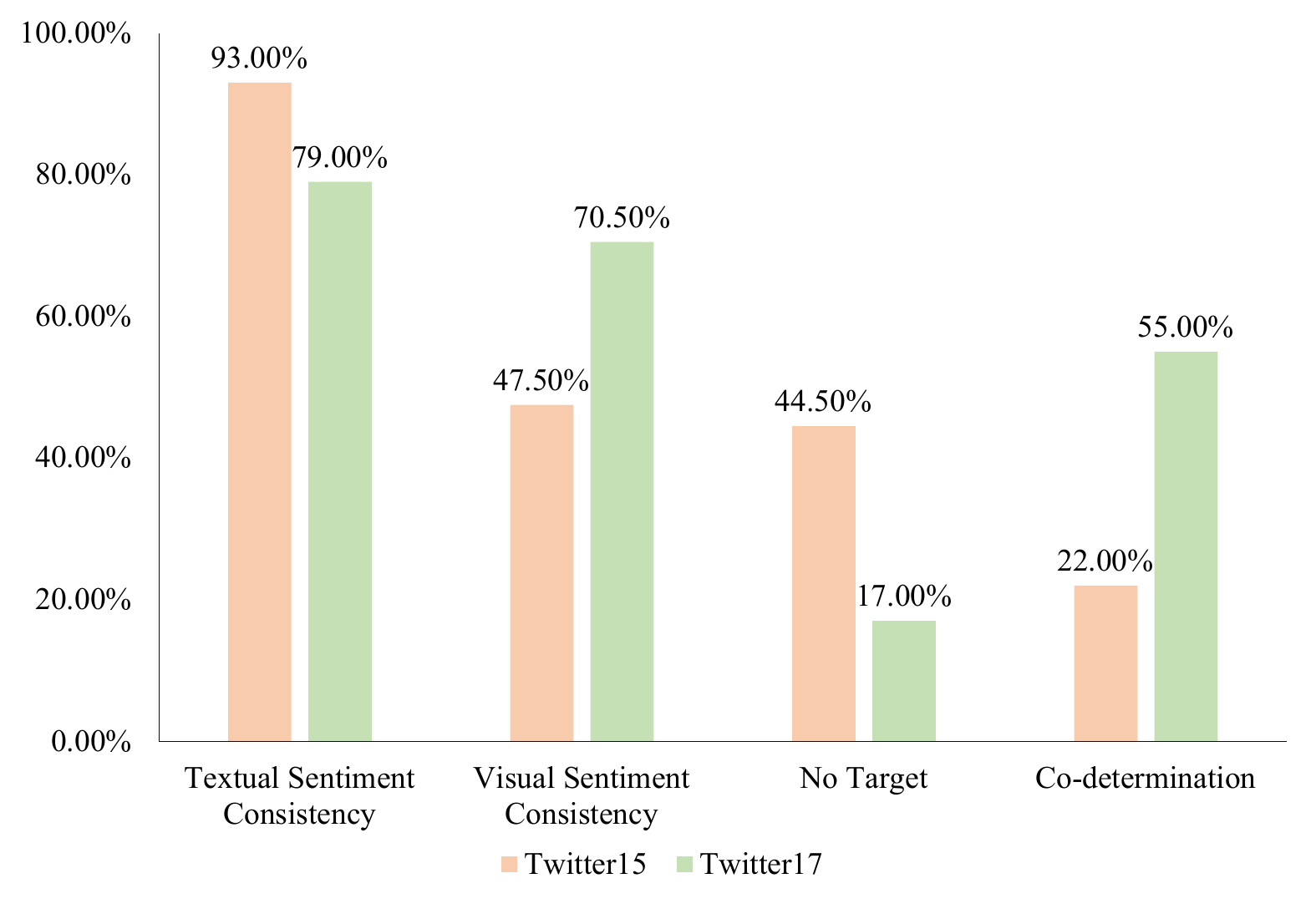}
\end{center}
\vspace{-4mm}
\caption{Annotation analysis.
Textual/Visual Sentiment Consistency: the consistency of the target's sentiment in text/image with the sentiment in multimodal information.  No Target: the percentage of images that are missing the target for sentiment analysis. Co-determination: the percentage of targets that sentiment is jointly determined by text and image.} 
\label{label consistency}
\vspace{-4mm}
\end{figure}


 To gain a deeper understanding of the performance issues mentioned above, 
 {we conduct detailed analyses of the two datasets, taking into account \emph{quantity}, \emph{diversity}, and \emph{annotation}. Following the annotation procedure employed by}~\citet{yu2019adapting}, {we enlist the participation of three domain experts to annotate 400 randomly sampled test data (200 from Twitter15 and 200 from Twitter17) across four aspects, with the majority vote being considered as the final annotation result} (Figure~\ref{label consistency}) \footnote{Illustrative examples with annotations are in Appendix~\ref{sec:case}.}.
We have the following observations:

\textbf{First}, as shown in Table~\ref{dataset}, the sample size is relatively small, with an average of less than 1.5 targets per sample.
Additionally, the distributions of the sentimental labels are unbalanced in both datasets, with neutral sentiment accounting for approximately 50\% and negative sentiment accounting for less than 15\%.
The reason behind this is that Twitter15 and Twitter17 were originally constructed by \citet{DBLP:conf/aaai/0001FLH18} and \citet{DBLP:conf/acl/JiZCLN18} respectively for the named entity recognition task, rather than specifically for TMSC.

    
\textbf{Second}, the multimodal sentiment has high consistency with the textual sentiment but low consistency with the visual sentiment.
In Twitter15, 93\% of the targets have the same textual sentiment as the multimodal sentiment, while only 47.5\% have a visual sentiment that matches. 
This indicates the biased distribution existing in the dataset, i.e., the textual information is more useful for determining the multimodal sentiment. Although this phenomenon is mitigated in Twitter17, the textual information is still more consistent with the multimodal sentiment than the visual information. 
    
\textbf{Third}, a large number of targets do not exist in  images,
which is also not suitable for the \emph{target-oriented} multimodal sentiment classification task. 
This 
{phenomenon} may stem from the construction of the two datasets, where the targets are selected directly from the text, without taking into account the corresponding images \cite{yu2019adapting}.
    
\textbf{Fourth}, due to the facts of irrelevant images and non-existence of targets in images, there is only a small portion of the data where the sentiment is determined by both text and images. Specifically, only 22\% of Twitter15 and 55\% of Twitter17 data require both text and images for the sentiment classification. As for the multimodal task, these two datasets may not be the best-suited in this aspect. 


{Based on our analyses of existing datasets, we propose that high-quality TMSC datasets should possess the following characteristics: 1) accurately reflecting the real-world data distribution, including factors such as unbalanced label distribution, while also providing sufficient data samples for different cases; 2) large data diversity, i.e., various data types and domains, to facilitate valid testing for models' generalization capability and robustness; and 3) multi-dimensional annotation information, including both multimodal and unimodal sentiment, to enable thorough analysis of the model's ability to handle different data sources.}

\section{Conclusion and Future Work}

In this paper, we conduct a series of in-depth experiments for TMSC and data analysis of existing datasets.
{Our findings reveal that current multimodal models do not exhibit significant performance gains compared to text-only models on the TMSC task. This is largely attributed to the over-reliance on textual modality in existing datasets, while visual modality playing a comparatively less significant role. Based on our experimental analyses, we propose future directions for designing models for the TMSC task and for constructing more suitable datasets which better capture the multimodal nature of social media sentiments.}


\section*{Limitations}
Although we have conducted a series of experiments and data analysis for the TMSC task to the best of our ability, there are at least the following limitations to our work.
First, our data analysis was performed mainly for the currently publicly available English datasets Twitter15 and Twitter17,
neglecting the Chinese dataset Multi-ZOL, which has not been widely studied.
Second, although our analysis indicated some problems in using the currently dataset to measure the TMSC task, we did not construct a new and better dataset for use in academic studies.
We have included this task as one of our future works to be investigated.
{Third, in our experiments, we did not specifically compare the impact of different text encoding methods on the model performance. While we acknowledge that different text encoding methods may indeed have an impact, it is worth noting that BERT, being a well-established text encoding method, already performs adequately. And most existing models use BERT as the text encoder. Therefore, we focused our study on investigating image encoding methods and fusion modules, as we believe there is more room for improvement in these parts.}




\section*{Acknowledgements}
The authors wish to thank the anonymous reviewers for their helpful comments. This work was partially funded by National Natural Science Foundation of China (No.62206057,61976056,62076069), Shanghai Rising-Star Program (23QA1400200), Natural Science Foundation of Shanghai (23ZR1403500), Program of Shanghai Academic Research Leader under grant 22XD1401100.

\bibliography{custom}
\bibliographystyle{acl_natbib}


\appendix

\section{Related Work}
As one of the tasks of sentiment analysis, TMSC has gained great attention from scholars in recent years~\cite{yu2019adapting}.
\citet{DBLP:conf/aaai/XuMC19} constructed a Chinese dataset named Multi-ZOL and proposed a multi-hop memory network for handling modal interactions.
Subsequently, \citet{yu2019adapting} constructed two English datasets, Twitter15 and Twitter17, and applied BERT to this task.
The following research on the TMSC task can be divided into two directions. On the one hand, there is the continuous exploration of how to enhance the interactions between modalities~\cite{2021Exploiting}, and on the other hand, there is the application of pre-trained models to this task~\cite{ye2022sentiment, DBLP:conf/acl/LingYX22}. Despite these efforts, the current models have not yet achieved significant performance gains relative to the text-only models. We have conducted a series of experiments and data analysis, hoping to provide some insights for the future research of TMSC.

\begin{table*}[t!]
\centering
\resizebox{\linewidth}{!}{
\begin{tabular}{m{3.8cm}|c|c|ccc}
\toprule
    \multirow{2}*{Text} \centering & \multirow{2}*{Image} & \multirow{2}*{Target} & \multicolumn{3}{c}{Sentiment}\\
    ~ & ~& ~& Multimodal & Textual & Visual  \\ \midrule
   Congratulations to our second draw winner - \textbf{Bulaire Leber} of ADSS Global , Haiti . Thanks for participating , Bulaire
&
\begin{minipage}[h]{0.4\columnwidth}
		\centering
		\raisebox{-0.5\height}{\includegraphics[width=\linewidth]{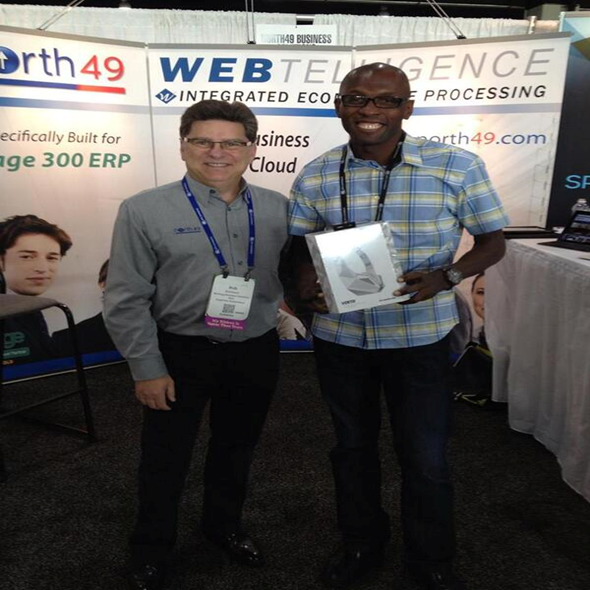}}
	\end{minipage}
 & Bulaire Leber & Positive & Positive \Checkmark& Positive \Checkmark\\ \midrule
 RT @ BeschlossDC : Coretta Scott King with Robert amp \textbf{Ethel Kennedy} after husband ' s assassination , which occurred tonight 1968
&
\begin{minipage}[h]{0.4\columnwidth}
		\centering
		\raisebox{-0.5\height}{\includegraphics[width=\linewidth]{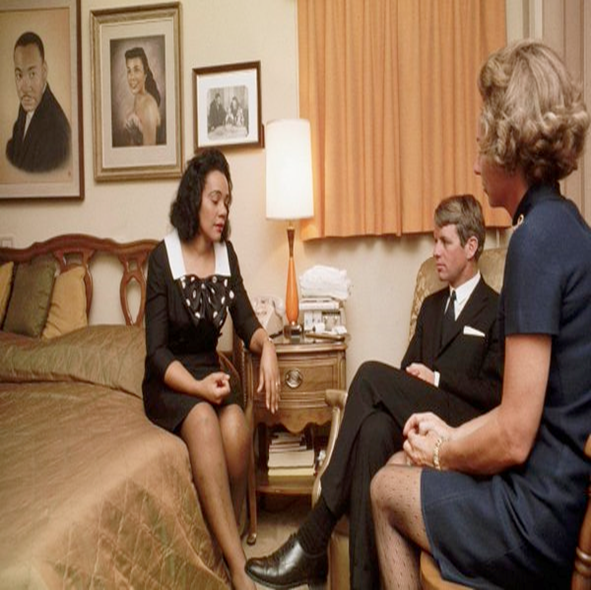}}
\end{minipage}
 & Ethel Kennedy & Negative & Negative \Checkmark & Neutral \XSolidBrush\\ \midrule Pres \textbf{Obama} takes the stage at @ RutgersU Commencement in school football stadium in Piscataway , NJ .
&     
\begin{minipage}[h]{0.4\columnwidth}
		\centering
		\raisebox{-0.5\height}{\includegraphics[width=\linewidth]{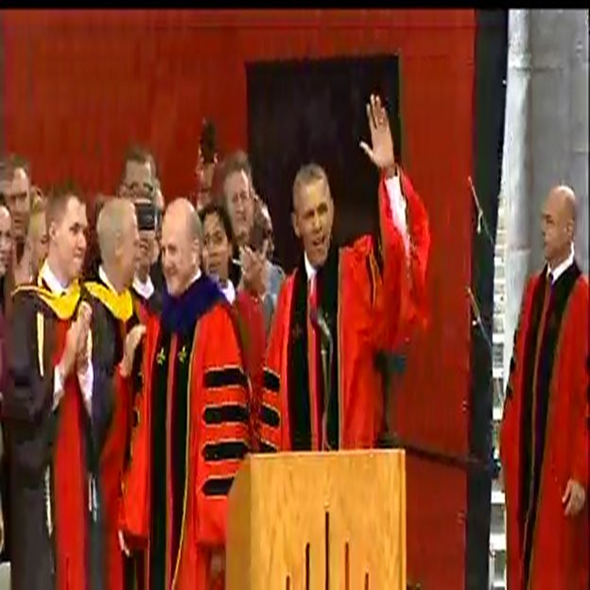}}
\end{minipage}
 & Obama & Positive & Neutral \XSolidBrush & Positive \Checkmark\\\midrule
 RT @ Refugees : Today , 18 - year - old Yehya became the 1 millionth Syrian to register as a refugee in \textbf{Lebanon}
&
\begin{minipage}[h]{0.4\columnwidth}
		\centering
		\raisebox{-0.5\height}{\includegraphics[width=\linewidth]{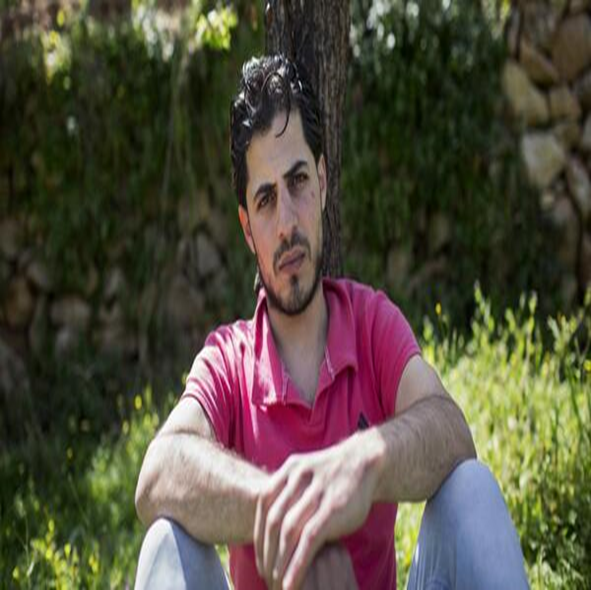}}
\end{minipage}
 & Lebanon & Neutral & Neutral \Checkmark & \underline{No Target} \XSolidBrush\\ 
\bottomrule
\end{tabular}}
\caption{The annotation examples.}
\label{tab:examples}
    
\end{table*}

\section{Experimental Setup}
\label{setup}

For each set of experiments, we conduct tests using five different random seeds (i.e., 0, 42, 199, 2022, and 11122). We initialize the parameters of the BERT text encoder with bert-base-uncased. The image encoder parameters are frozen, and we use resnet-152, vit-base, and faster-rcnn retrained on the Visual Genome dataset as image encoders, respectively. For both the self-attention and cross-attention modules, we use the last three initialization parameters of bert-base-uncased. We utilize the Adam optimizer~\cite{adam} with a learning rate of 2e-5 and run each experiment on a 3090 GPU for 8 epochs. We select the best epoch's parameters on the validation set for testing and calculate the mean and standard deviation as the final result.

To ensure a fair comparison, we set the models in Table~\ref{experiment} uniformly and without continuing pre-training, which may make it challenging to compare them with existing papers due to differences in overall structure and training details compared to Table~\ref{baseline}, even if they may use the same unimodal encoder and multimodal fusion modules.

\begin{figure}[t!]
    \centering
    \subfigure[Twitter15]{
    \label{venn15}
    \includegraphics[width=1.4in]{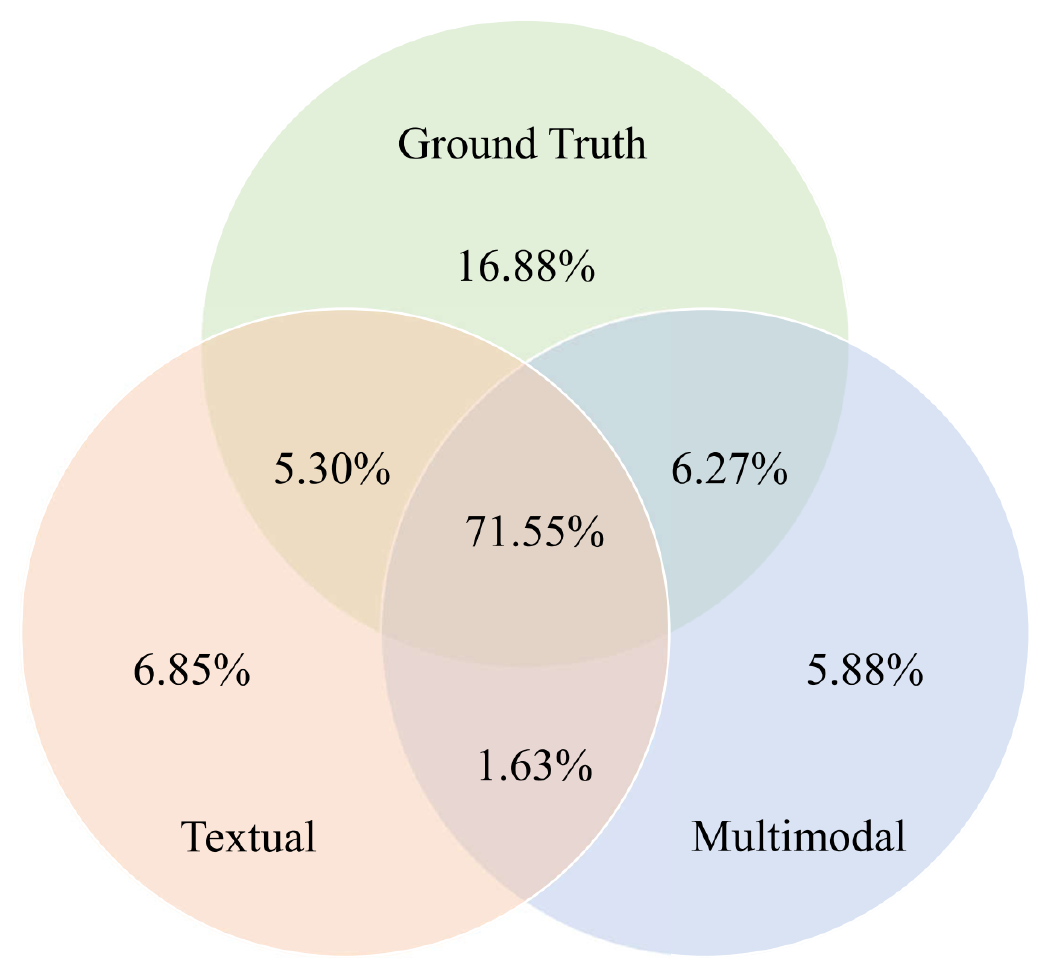}
    }
    \subfigure[Twitter17]{
    \label{venn17}
    \includegraphics[width=1.4in]{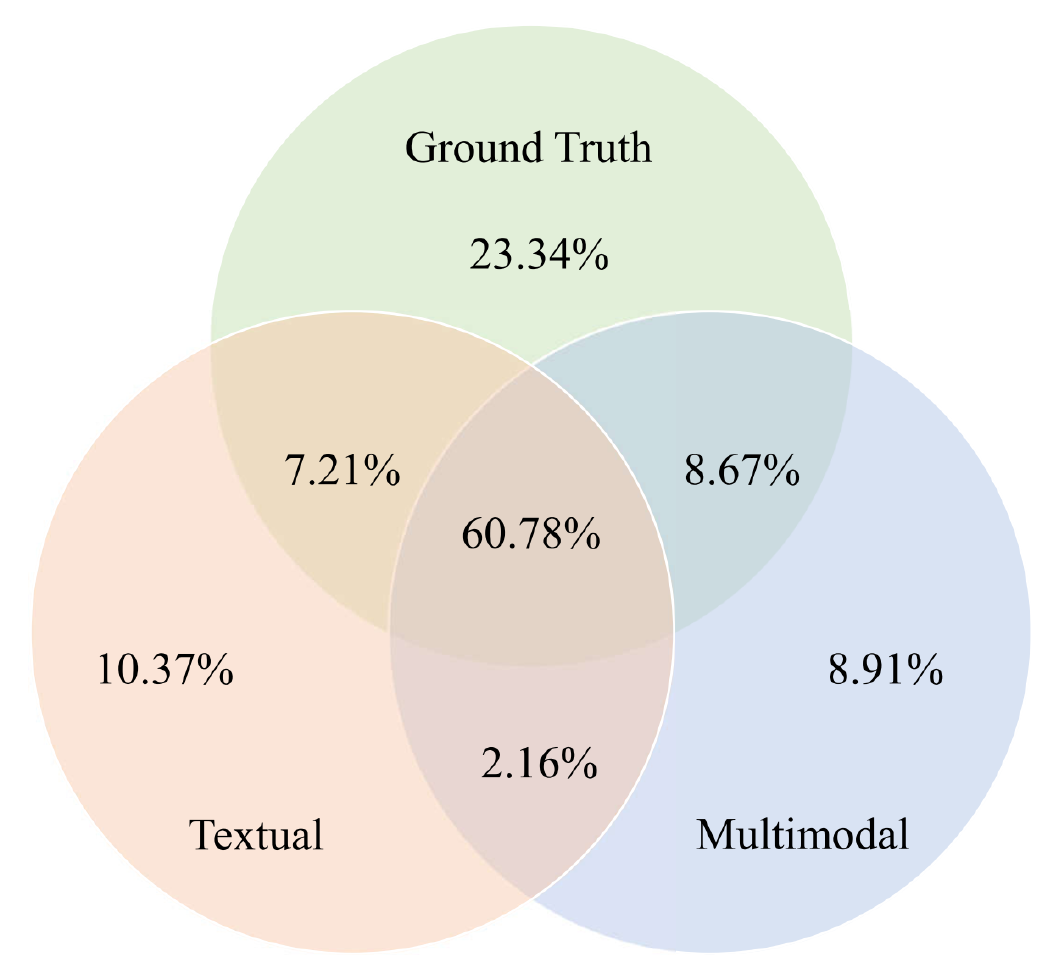}
    }
    \caption{Venn diagram for model performance visualization.}
    \label{fig:venn}
\end{figure}

\section{Model Performance Visualization}
\label{visualization}

We select Image2Text (Faster R-CNN) as the representative of the multimodal models and compare its performance with that of BERT with a random seed of 11122 to obtain Figure~\ref{fig:venn}.
The intersection of every two circles in the figure represents the part where the prediction results are consistent.
Based on this comparison, we have the following observations:
\textbf{First}, in terms of prediction accuracy, the multimodal model does not achieve a significant gain over the text-only model.
\textbf{Second}, a portion of the data is predicted correctly by the multimodal model but incorrectly by the text-only model, and vice versa. The proportions of these two parts are similar. This suggests that when images do contribute valuable information to the multimodal model, they also introduce noise. In order to improve the performance, further investigation is required for how to properly incorporate the visual information. 
\textbf{Third}, over 16\% of the data has sentiments that neither the text-only model nor the multimodal model predicts correctly. This indicates the weakness of the current models and we need further explorations. 

\section{Annotation Examples}
\label{sec:case}

To clearly and visually illustrate the various scenarios that arise during the dataset annotation process, four samples are presented in Table~\ref{tab:examples}.

The \textbf{first} example demonstrates a scenario where the textual sentiment and the visual sentiment matches, resulting in a multimodal sentiment determined by both modalities.
In the example, the sentiment in the text is determined to be positive through the use of words such as ``Congratulations'' and ``winner''. Similarly, the sentiment in the image can be inferred as positive by identifying the target (i.e., the first person on the right) and noticing his smiling face.

The \textbf{second} example shows a scenario where the textual sentiment aligns with the multimodal sentiment but not with the visual sentiment, leading to a multimodal sentiment determined by the textual modality only. 
Specifically, the sentiment conveyed by the text is negative due to the phrase ``after husband's assassination'' and the sentiment conveyed by the image is neutral as it does not show an obvious facial expression on the person referred to in the text (i.e., the first person on the left). Therefore, the multimodal sentiment conveyed by both modalities is negative.

Corresponding to the second example, the \textbf{third} example illustrates a scenario where the visual sentiment aligns with the multimodal sentiment but not with the textual sentiment. 
In particular, the text simply states a fact with a neutral sentiment, while the image shows the target (i.e., the person waving his hand in front of the podium) with a positive facial expression and posture, resulting in a positive multimodal sentiment overall.

The \textbf{fourth} example presents a scenario where there is no target in the image, resulting in a multimodal sentiment determined solely by the textual modality. 
Here, the target is ``Lebanon'', but since there is only one person in the image and no information about ``Lebanon'', we can only conclude that the multimodal sentiment is neutral based on the text.
It is worth mentioning that such a sample is not ideal for the TMSC task as the image does not convey any sentimental information towards the target. 


\end{document}